\documentclass[10pt,letterpaper]{article}

\usepackage{cogsci}
\cogscifinalcopy 

\makeatletter
\renewcommand*{\@fnsymbol}[1]{\ifcase#1\or*\else\@arabic{\numexpr#1-1\relax}\fi}
\makeatother

\usepackage{pslatex}
\usepackage{apacite}
\usepackage{graphicx}
\usepackage{array}
\usepackage{multirow}
\usepackage{booktabs}
\usepackage{tabularx}
\usepackage{amsmath}
\usepackage{amssymb}
\usepackage{float}
\usepackage{subcaption}
\usepackage{numprint}
\usepackage[bottom]{footmisc}

\usepackage{float} 

\graphicspath{{./fig/}}

\title{Learning to cooperate: Emergent communication in multi-agent navigation}
 
\author{
    {\large \bf Ivana Kaji\'c\thanks{Research done during an internship at DeepMind.}}  (ivana.kajic@uwaterloo.ca)\\ 
    Centre for Theoretical Neuroscience, University of Waterloo  \\
    Waterloo, ON, Canada\\
    \AND
    {\large \bf Eser Ayg\"un (eser@google.com)}\\
    {\large \bf Doina Precup (doinap@google.com)} \\ DeepMind, Montr\'eal, QC, Canada \\
}

\begin{document}
\npthousandsep{,}  

\maketitle

\begin{abstract}
Emergent communication in artificial agents has been studied to understand language evolution,
as well as to develop artificial systems that learn to communicate with humans.
We show that agents performing a cooperative navigation task in various gridworld environments learn an interpretable communication protocol that enables them to efficiently, and in many cases, optimally, solve the task.
An analysis of the agents' policies reveals that emergent signals spatially cluster the state space, with signals referring to specific locations and spatial directions such as \emph{left}, \emph{up}, or \emph{upper left room}.
Using populations of agents, we show that the emergent protocol has basic compositional structure, thus exhibiting a core property of natural language.

\end{abstract}

\section{Introduction}
The remarkable diversity of the world's languages is attributed to a variety of
factors including socio-cultural conventions among users of a language,
physical environments, and neurocognitive mechanisms~\cite{lupyan2016}. Humans also posses
a striking ability to adapt their communication in light of changes to
such factors, for example, by coordinating their actions to rapidly develop novel communication systems that exhibit core features of natural language, such as referential signaling and compositional structure~\cite{bohn2019}.

The extent to which such communicative adaptability and linguistic
structure are present in communication systems learned by artificial
intelligence remains an ongoing topic of research. In particular, interactive
multi-agent setups in reinforcement learning, where communicating agents learn
to reason about novel signals while being situated in environments~\cite{li2019,lazaridou2018,mordatch2018,kottur2017} are   
a promising approach for the study of emergent languages and their relationship to natural language.

Often, such setups study emergent languages in the context of 
cooperative games, where agents with assigned roles need to coordinate their
actions to achieve a goal of mutual interest. For example, the signaling game of~\citeA{lewis69} is used, where the sender sees an artifact (e.g., an image) and uses
a communication channel shared with a receiver to transmit a message (e.g.,
a symbol) from a fixed vocabulary.
The receiver then conditions its decision on the message to select the target object among distractors.
Communication protocols in such games exhibit some level of structure
reminiscent of natural language. In
particular, compositional structure is found with end-to-end training on disentangled input
data~\cite{lazaridou2018} or by introducing environmental pressures during
training~\cite{li2019}.

While referential games provide a useful paradigm for investigating conditions
that give rise to communication protocols with natural language-like
properties~\cite{kottur2017}, the resulting communication policies are often
difficult to interpret~\cite{lowe2019}, and agents' actions in such setups
often do not affect the environment, as the sender and receiver are each
restricted to one choice of action.

Here, we contribute to this body of research by extending the
referential game as a navigation task.
We introduce a control component where an agent needs to reach
a specific, to it unknown, location in the environment by relying on signals received from another agent which knows the location.
This extension of the task is akin to a situation in which a tourist asks
a stranger on the street for route directions to a particular location in
a city they are visiting for the first time. Such directions are likely to contain the description
along the lines of ``Go up this street, take the 2nd street to the right, then
1st to the left''.
We show that the communication protocol learned by our agents exhibits features
that can be interpreted in a similar way, while being optimal for this task, as demonstrated by agents' performance on the task.
In addition, we demonstrate that when a population of agents coordinates their joint description of a route, even without knowing each other's descriptions, the protocol exhibits basic compositional structure, where individual messages encode different spatial aspects of the environment such as \emph{left}, \emph{up}, or \emph{upper left room}.

\begin{figure*}[t!]
\includegraphics[scale=0.7]{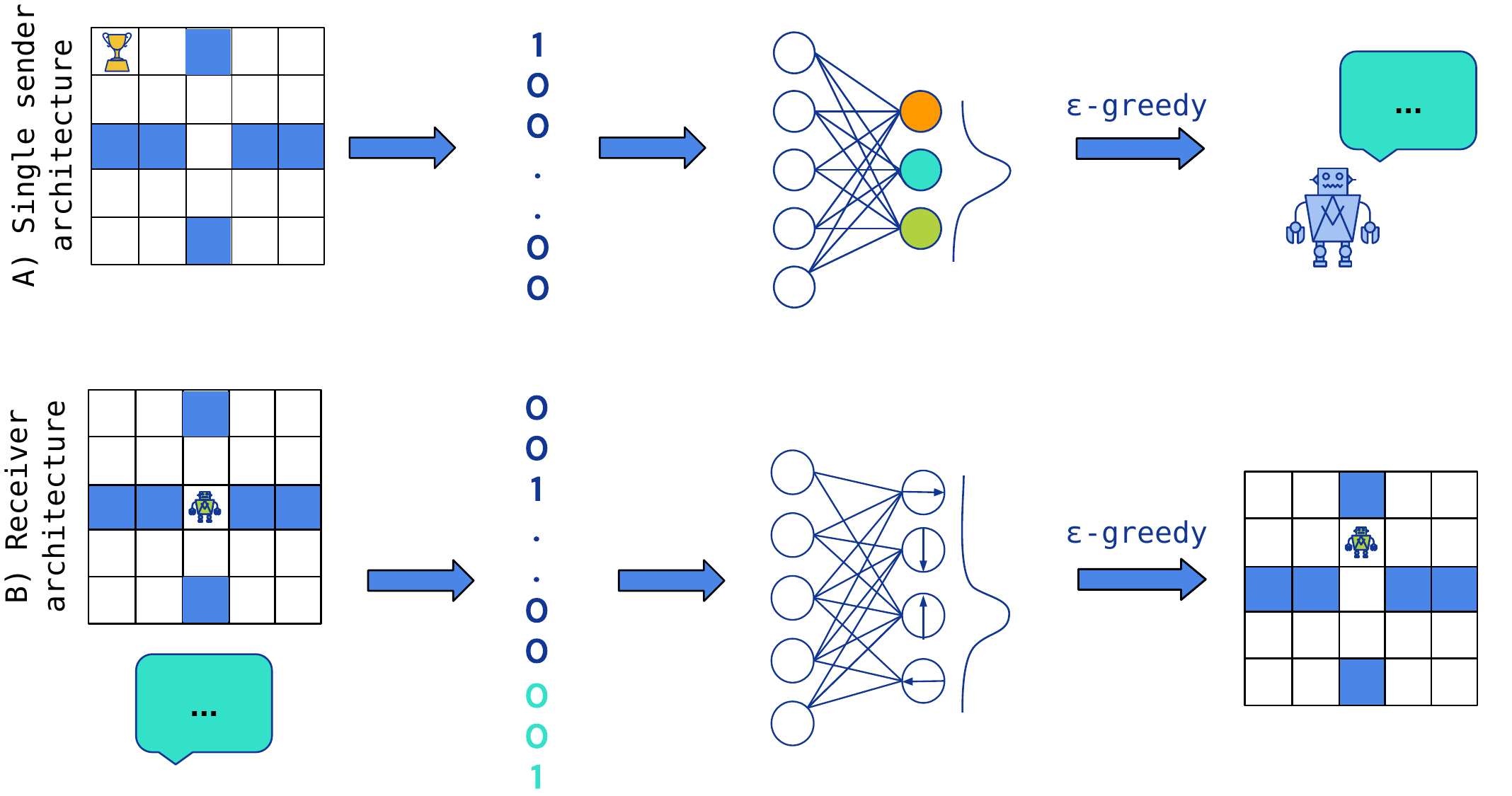} 
\centering
\caption{
    Both the sender and the receiver see the gridworld environment, yet
    only the sender sees the goal location (A). It selects a message action (a single
    symbol) based on the one-hot encoding of the goal location.
	The receiver selects a navigation action based on the multi-hot input vector
	that encodes its own location and the message (B).}
    \label{fig:agents}
\end{figure*}

We investigate the relationship between the environment topology and signals for the description of paths in an environment using a simple setup with minimal or no assumptions on rewards or the protocol.
We perform a qualitative analysis of the protocol and discuss it the context of different linguistic strategies used by people describing navigation in different  environments.

\section{The Navigation Task}
We consider a cooperative navigation task, where one agent (the sender) sees the goal location in a gridworld environment, sends a message to another agent (the receiver) who has to reach that location. Both agents are rewarded when the receiver reaches the goal location.
In a given task setup, there is exactly one receiver, and at most five senders.
At the beginning of each episode of the task ($t=0$), the goal location is determined 
by placing a reward at a random, non-occupied location in the gridworld environment.
The sender observes the goal location and in response, it emits
a message $m$ from a vocabulary $V$, with $|V|=N$.
The sender does not see the receiver location.
All messages are represented as single discrete symbols (here, we use natural
numbers as symbols).
The message is stored by the environment and provided as part of the
observation available to the receiver. The message observation persists at each subsequent time step $t>0$.
The sender performs no more actions until the end of the episode.

After $t=0$, the receiver observes its own location (center of the map) and the message, and performs a navigation action.
This process repeats until either the goal state is reached or the episode terminates, with  probability $p_{term}$.
If the episode terminates because the receiver reached the goal state, both receiver and sender are rewarded with $r=1$, and $r=0$ otherwise.

If the setup involves a population of senders, the interaction remains as
described, except that the message emitted at $t=0$
consists of a sequence of symbols $[m_1, m_2, ..., m_M]$, where $m_i$ is the
message emitted by the $i$-th sender and $M$ is the number of senders.
A sender selects its own message action independently of the messages selected by other senders, that is, one sender does not know the message emitted by other senders.
We use this multi-sender setup to investigate the properties of a set of messages encoding space, and in contrast to some other approaches (e.g., using RNNs to encode a sequence of messages), this approach imposes no assumptions on the protocol structure.

\subsection{Experimental setup}
\paragraph{Sender}
A sender agent is modelled as a contextual $N$-armed bandit that selects one
message action\footnote{For the sender, we interchangeably use the terms ``message'' and ``message
action''.} $m$ out of $N$ possible messages, based on the static context vector $c \in \mathbb{R}^d$,
where $d$ is the size of the gridworld environment.
The context is a one-hot vector encoding the goal location in the flattened array
representing the gridworld. In the experiments, we varied  $N$ to study the effects of more compression on the task performance.

The $i$-th sender's action-value estimation function $Q(\cdot)$ is implemented  as a single layer feed-forward neural network parameterized by
$\theta_{s_i}$. The loss for a single sender is 
$ \mathcal{L}_{s_i} =\big( R_t - Q(c, m_i; \theta_{s_i}) \big)^2$ when $t=T$,
and $0$ when $ 0 \leq t <  T$, where $R_t$ is the reward received at the end of an episode of length $T$ with the value of 1 or 0 depending on whether the receiver reached the goal state,
$c$ is the context
and $m_i$ is a message action. 
Message actions are selected using an $\epsilon$-greedy policy, where
$\epsilon$ is the same for all senders and is determined empirically using hyperparameter search.
The implementation of a sender agent is schematically shown in Figure~\ref{fig:agents}A.
\begin{table*}[t!]
    \centering
    \caption{Hyperparameters and settings used to train different sender-receiver agent configurations. Hyperparameters in italics are from experiments reported in the paper.}
    \begin{tabularx}{\textwidth}{cXX}  
     \toprule
Name & Values & Description \\ \midrule
$M$       & $[1, 2, 3, 4, 5]$ & Number of sender agents \\
$C$       & $[3, 4, 5, 8, 9, 16, 25, 27, 32, 36, 64]$ & Communication channel
capacity ($N^M$) \\
$\eta$    & $[5\text{e-}5, \textit{1\text{e-}4}, \textit{5\text{e-}4}, \textit{1\text{e-}3}]$ & Learning rate for RMSprop\\
$\epsilon_s$ & $[\textit{0.01}, \textit{0.05}, 0.1, 0.15]$ & Sender's action exploration rate \\
$\epsilon_r$ & $[\textit{0.01}, \textit{0.05}, \textit{0.1}, 0.15]$ & Receiver's action exploration
rate \\
$\gamma$ & $[0.7, 0.8, 0.9]$ & Receiver's Q-learning discount factor \\
$layout$ & [Pong, Four room, Two room, Flower, Empty room] & Environments
(see Fig. \ref{fig:envs}) \\
\bottomrule
\end{tabularx} 

    \label{tab:hypers}
\end{table*}

\paragraph{Receiver} The receiver is implemented as a Q-learning agent~\cite{watkins92} with
a neural network representing action-values and parameterized by $\theta_{r}$. After
$t=0$, the environment provides the receiver with an observation $o_t=[p_t,
\bar m_1, \bar m_2, ...]$ at each time step $t$, where $p_t$ is a one-hot encoding of the receiver position, and each $\bar m_i$ is a one-hot encoding of the message emitted
by sender $i$. $o_t$ is provided as the input to the neural network, and
outputs are Q-values for each of the four possible navigation actions (up,
down, left and right). An action is selected using an $\epsilon$-greedy
policy, and the temporal difference~\cite{sutton87} error is used to compute the learning loss, i.e.:
$ \mathcal{L}_{r} = \big( R_t + \gamma \max_{a} Q(o_{t+1}, a; \theta_{r})
- Q(o_t, a_t; \theta_{r}) \big)^2,  $ when  $0 < t \leq T$ and $0$ at $t=0$.

In order to additionally incentivize the agents to adopt efficient behaviours, such
as reaching the goal state using the shortest possible path, the random
termination probability is set as $p_{term}=1-\gamma$, where $\gamma$ is the discount factor in
the Q-learning algorithm. The total loss $ \mathcal{L}_{total} = \Sigma_i
\mathcal{L}_{s_i} (\theta_{s_i})+ \mathcal{L}_{r}  ( \theta_{r})$ is then
minimized using the RMSProp optimizer with a minibatch size 10, implemented in TensorFlow.

\begin{figure}[ht]
\includegraphics[scale=0.44]{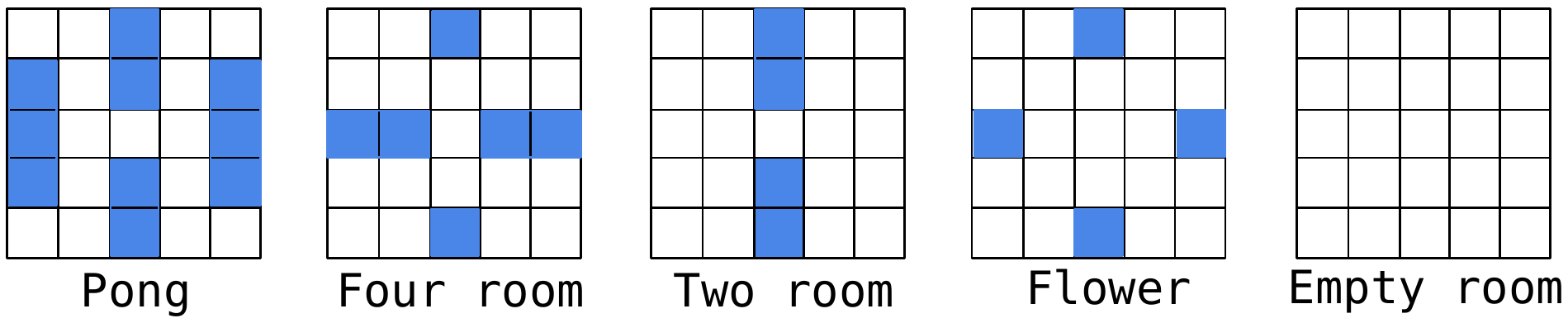} 
\centering
    \caption{The five different gridworld environments used in experiments. Blue cells represent walls.} 
\label{fig:envs}
\end{figure}

\paragraph{Environment}
The receiver and the reward are situated in one out of five possible gridworld environments shown in Figure~\ref{fig:envs}.
All layouts have the same dimensions (5$\times$5), and they differ in the placement of walls.
These layout were selected to investigate the influence of the environment  structure on the emergent communication signals.
The environment is non-deterministic due to $p_{term}$, as the episode may end randomly in each step for the receiver, excluding the first step.

\paragraph{Training}
Each experiment consisting of a sender-receiver agent setup is trained for 20 million steps in a gridworld environment. 
Each step is a step within a single episode, and the length of the episode is determined by the termination probability $p_{term}$, so that on average each episode consists of 3, 5 or 10 steps. Shorter episodes impose an additional pressure on agents to act efficiently.
Table~\ref{tab:hypers} lists hyperparameter values and descriptions across experiments.
In experiments that contain more than one sender, all senders have the same
$\epsilon$ value, and the same vocabulary size $N$.
The number of available messages depends on the number of senders, and is
selected so as to allow a fair comparison among different
configurations in terms of the total number of messages (more details are
provided in the following sections).
In the analyses, we select the runs with the best performing learning rates and
exploration rates, resulting in approximately 180k experiments.

\section{Results}
\begin{figure*}[t!]
\includegraphics[scale=0.491]{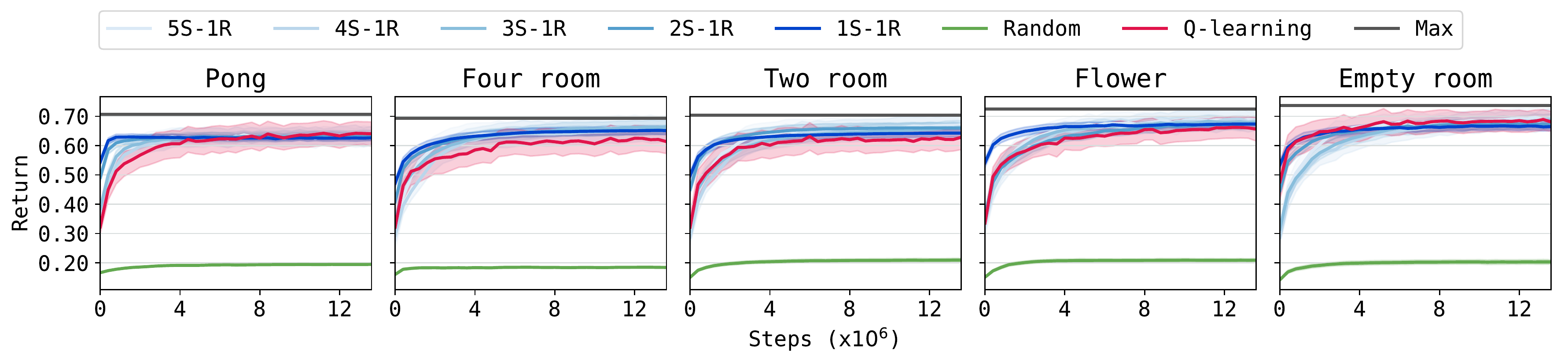}

\vspace{.5cm}
\includegraphics[scale=0.492]{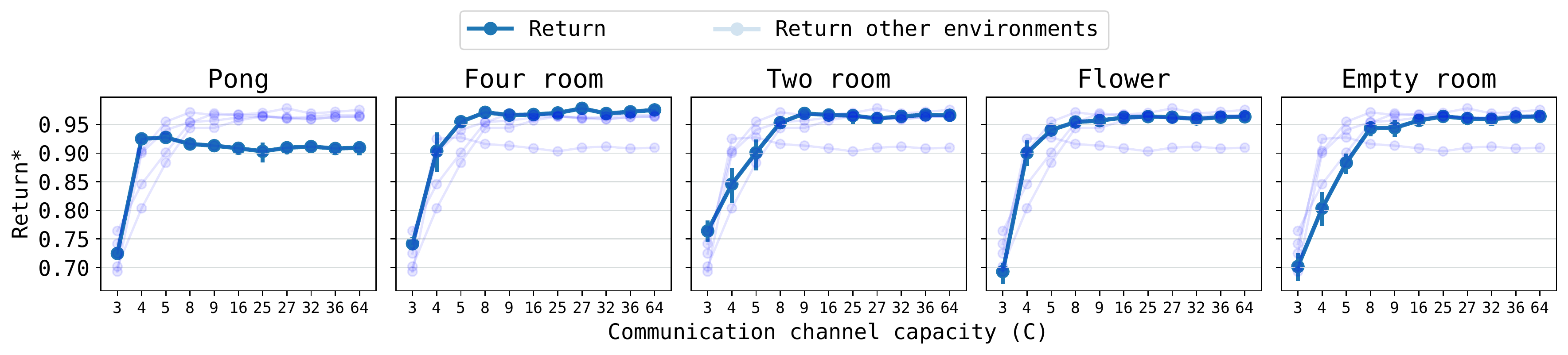}
    \caption{
    Upper panel: Mean training return curves for different sender-receiver agent
    configurations with baseline comparisons (Q-learning, Random and Theoretical maximum).
    Lower panel:
    Increase in average return depending on the channel capacity $C$,
    defined as the total number of messages $N^M$.
    Abbreviations: S=Sender(s), R=Receiver. Return*=Return normalized by theoretical maximum.}
    \label{fig:return}
\end{figure*}
To evaluate learning success, we first analyze the average returns obtained during training.
The upper panel in Figure~\ref{fig:return} shows the returns sampled at regular intervals during training for different
sender-receiver configurations, for the first 12 million steps.
The highest possible return for each environment is indicated by a gray line,
and is calculated as the average of all possible discounted rewards in the given
environment, when using the optimal policy (i.e. the shortest path).
The figure also shows comparisons with two additional baselines: a single,
non-communicating Q-learning agent that sees the goal, as well as a random baseline, which consists of a sender emitting random messages. In the latter case, the receiver's optimal strategy is to treat the messages as noise, and learn to visit
every possible location in the environment in search for the goal.

Based on the training curves, we make two major observations. First,
communicating agents are able to successfully solve this task, as shown by the
learning curve approaching the theoretical maximum value. The performance of
communicating agents is comparable to that of a single \mbox{Q-learning} agent.
Second, we observe that different combinations of communicating
agents, and in particular the configuration consisting of a single sender and
a single receiver, often display faster convergence than the single Q-learning
agent. This is apparent in all environments, except in the \emph{Empty room}
environment where they are comparable. We note that this could also be due to
the lack of extensive tuning of neural network hyperparameters, such as the number of units in the hidden layer or the regularization
factors. From our experiments, it appears that the single Q-learning agent was
more affected by the lack of hyperparameter tuning, but our goal was not to examine the conditions under which the agents perform optimally on this task.
Instead, we are interested in the coordination problem when two or more agents need to cooperate to solve the task, given individual restrictions on the action space. The task is purposefully designed to be sufficiently simple to yield an interpretable solution to the coordination problem that allows us to understand the influence of  different environmental pressures on communicative success.

Finally, we note two reasons why the training curves do not reach the theoretical maximum. First, these were generated in the training regime which contains some level of random behaviour as determined by the $\epsilon$-greedy strategy.
Second, these curves are averaged over all communication channel capacities, and for some of them the task cannot be solved optimally with the Q-learning algorithm (discussed later).

\subsection{Communication channel capacity}
Next, we study how the capacity of the communication channel affects the
agents' performance on the task. The capacity $C$ of the communication channel
is defined as the total number of messages used in an agent setup, and is computed as $C=N^M$.
For example, $C=16$ corresponds to the following setups:
1 sender with 16 messages, 2 senders with 4 messages each, or
4 senders with 2 messages each.
We examine the relationship between the channel capacity and the average return
normalized by the theoretical maximum.

We know that the agents should perform well on the task if the size of the
communication channel is the same as the number of non-occupied locations in the
environment. In this case, every message can be used to uniquely identify one
goal location.
However, we are particularly interested in understanding the meaning of
messages when the size of the channel is small compared to the number of
possible goal locations.
For example, if there are only 3 or 4 messages available, the sender needs to
compress information about goal locations and use a single message to convey
information about several goal locations.

The lower panel in Figure~\ref{fig:return} shows normalized returns for different sizes of
communication channel, averaged over all agent setups.
A single point on a curve is an average return by all sender-receiver 
agent configurations that use the total number of messages  indicated by the corresponding
channel capacity $C$ on the \emph{x}-axis.
In all environments, we observe two distinct curve regions:  a linear increase in the performance with every message
added to the communication channel, and a plateau where adding more messages does not improve performance.
These regions are most apparent in the \emph{Empty room} and \emph{Two room}
environments, where we see an increase in performance up to approximately 8 or
9 messages.
For the \emph{Pong} environment, the peak is reached at about 4 messages, and
the performance drops when there are more than 5 messages available.

The point after which we observe little to no improvement is
different for each environment, and it approximately corresponds to the
smallest number of messages needed to encode shortest paths to all possible goal locations. 
For example, in the \emph{Pong} environment 4 such paths exist, so 
when the goal is located anywhere on one of the paths, the agent is guaranteed to reach the
goal using the fewest steps possible, if it knows which path to take.
For all other environments, there are 8 such paths.
Such path configurations may not be unique, as for some environments there are multiple shortest paths to each goal location.
While having 8 messages requires agents to compress information
about goal locations, this compression still allows them to optimally solve the task,
and can thus be seen as a form of lossless compression.

\subsubsection{Subcapacity and supracapacity regime}
The subcapacity regime is defined as 5 or fewer messages in total for
\emph{Pong}, and 9 or fewer messages for all other environments.
Approximately, these thresholds correspond to the channel capacity regions in which agents need to use lossy compression (subcapacity) or not (supracapacity).
In the subcapacity regime, increasing the capacity of the communication channel
by adding an additional message strongly correlates ($r>.65, p<.001$) with the improvement in performance for all environments.

We also investigate the relationship between the environment structure and the
normalized performance.
Here, we consider structure as a measure of the uniqueness of the optimal policies, and quantify it as the inverse of the total number of shortest paths to each location in the environment.
According to that definition, we obtain
the following values: 1/14 for \emph{Pong}, 1/22 for \emph{Four room}, 1/32 for
\emph{Two room}, 1/44 for \emph{Flower} and 1/64 for \emph{Empty room}.
Consequently, \emph{Pong} is the environment with most
structure, and \emph{Empty room} has the least structure.
This measure is also related to the amount of empty space in the
environment, as environments with more empty space are less structured.

We find that agent pairs in the subcapacity regime, specifically in the low regime of $C=3,4$ achieve higher normalized return in  environments with
more structure ($r=0.42$, $p<.001$).
In this case, the structure is helpful insofar as obstacles in the environment restrict possible paths for the receiver. 
This effect is reduced as more messages are added, and even reverses in the supracapacity regime.

\subsection{Emergent Communication Protocol}
We now focus on the analysis of the agents' policies in order to characterize the learned communication protocol.
To understand the information conveyed by the senders' messages, we manipulate goal
locations, and
examine what effect the manipulation has on the emitted messages.
The manipulation consists of placing the goal at all possible
locations in the environment.
We also analyze the receiver's policy by manipulating messages and observing the resulting trajectories in the environment.

In this section we will be predominantly concerned with the
following two questions:
\begin{enumerate}
    \item What message does the sender choose for goal 
         \emph{(x, y)}?
    \item Where does a receiver go if it receives a message \emph{m} while
         at location \emph{(x, y)}?
\end{enumerate}

\subsubsection{Qualitative analysis}
\begin{figure}[t]
\includegraphics[scale=0.18]{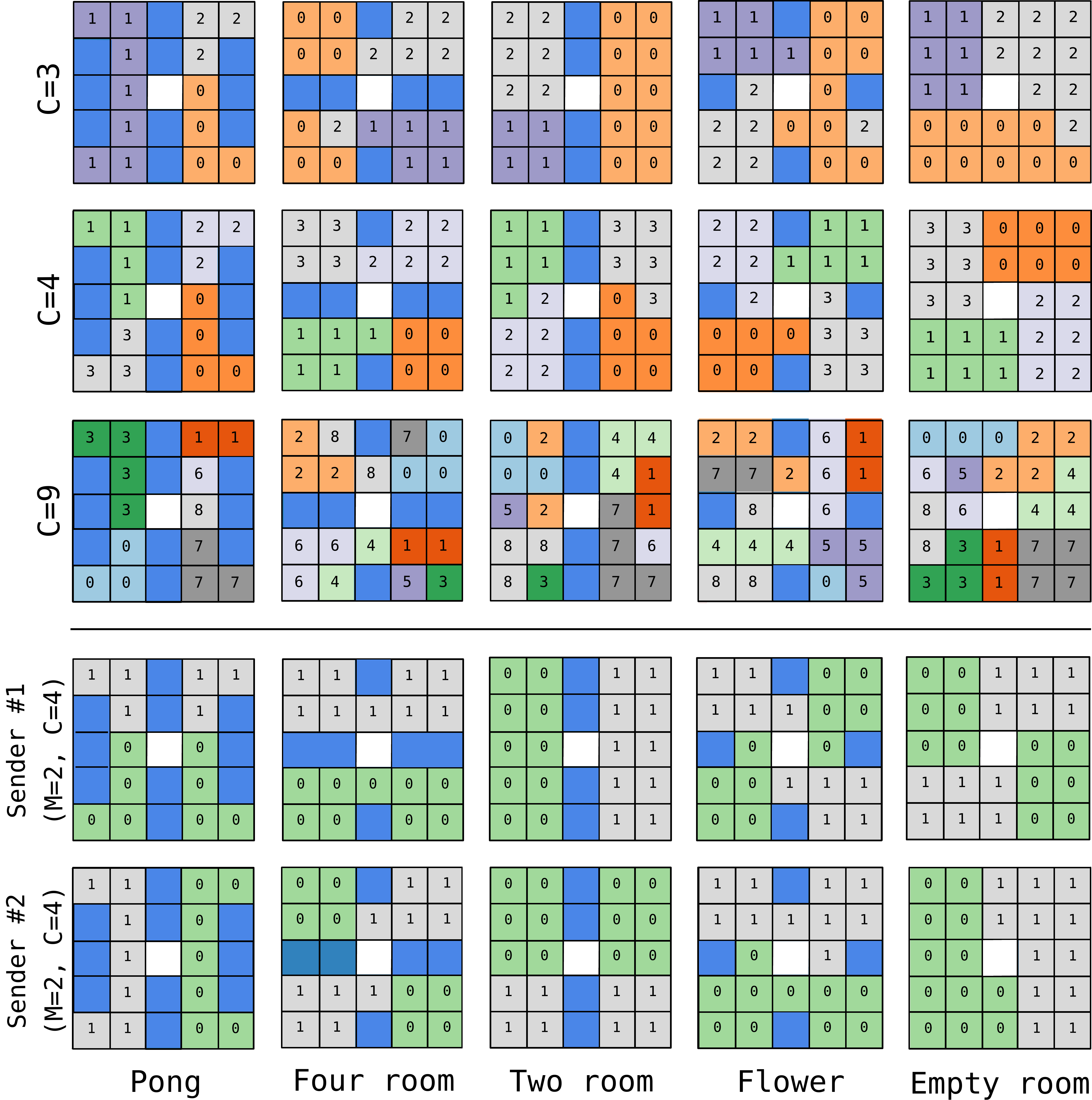}
\centering
\caption{Sender's policy: Learned message distributions for goal locations for different channel
    capacity sizes (C) and different agent setups (upper panel: 1S-1R, lower
    panel: 2S-1R).}
    \label{fig:messages}
\end{figure}

To answer the first question, we follow the procedure shown in
Figure~\ref{fig:agents}A). From a trained sender-receiver model we use the sender's network, and probe it with a one-hot encoded vector that represents a single goal location \emph{(x, y)}.
We then use greedy action selection ($\epsilon=0$) to select a message at the
output.
This process is repeated for all goal locations for a specific environment, obtaining a single message for each
location.
Then, locations with the same message are color-coded, yielding a single plot in
Figure~\ref{fig:messages}.

The upper panel contains message distributions from 15 different experiments, each one corresponding to a different 1S-1R setup in different environments and for channel capacity sizes of $C=3, 4, 9$. The
center, corresponding to the location $(2,2)$, is left blank in all plots, as
the starting position of the receiver can never be a goal location.

Message distributions generally produce clustered regions in
space, so that goal locations anywhere in a region are described by a single message.
The effects of lossy compression are apparent when $C=3, 4$, as in such cases a single message is assigned to multiple goal locations.
Such regions are also interpretable: for example, when $C=3$ in the
\emph{Pong} environment, a single message encodes \emph{left of the center}, and in the \emph{Two room} environment, a single message encodes \emph{right of the center}.
$C=3$ is a particularly interesting case for a Q-learning agent in some environments, as the agent bases its next action only on its current location and the message.
As such, it cannot learn a strategy where it ``goes back'' to a previously visited location and select a different action in that state.
For example, once the agent enters the left corridor in \emph{Pong} ($C=3$) upon seeing the message ``1'', it learns to navigate either up or down in that corridor. It is only due to random action selection that it visits the other part of that corridor.
Due to this restriction, agents in this setup can never perform optimally in a sense that they will never use the shortest possible path to all possible goal locations.

\begin{figure}[t]
\includegraphics[scale=0.135]{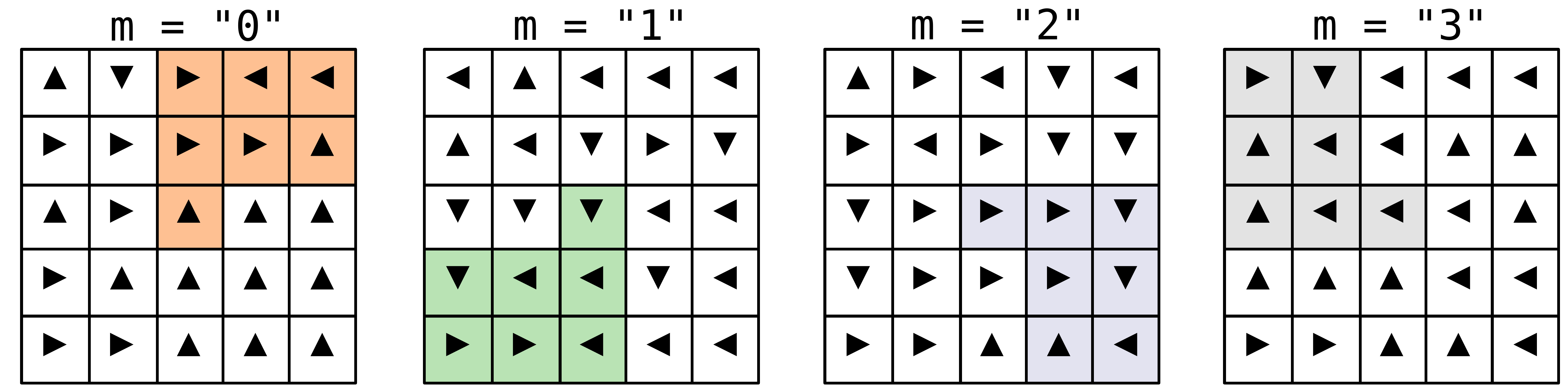}
\centering
\caption{
    Receiver's policy: Navigation trajectories as actions with the highest
Q-value for each location and each message $m$ for \emph{Empty room} $(C=4)$.
    }
    \label{fig:flow}
\end{figure}

Finer spatial interpretations emerge with $C=4$, where message clusters 
encode regions that can be described as individual rooms, such as those in
\emph{Four room}, or equal and symmetrical areas of space, as those in
\emph{Flower} $(C=4)$ or \emph{Empty room} $(C=4)$.
With this kind of compression taking place, we often observe that
the receiver adopts a ``sweeping'' goal search strategy.
Figure~\ref{fig:flow} shows the receiver's policy for the sender's policy \emph{Empty
room} $(C=4)$ shown in Figure~\ref{fig:messages}.
For each possible message, the receiver's greedy actions for each location \emph{(x, y)} are depicted as arrows.
That is, a single arrow at a location provides an answer to the question
2 above.
Iteratively answering this question yields a trajectory through the space, starting from the center.

In this example, all trajectories cover every location in their highlighted region of space only once, before looping back to a previously
visited location. Assuming the goal is located on the trajectory, the path
defined by that trajectory is Hamiltonian and is thus an optimal navigation
strategy for this capacity regime.

As we increase the size of the capacity channel, we observe that messages start
encoding shortest paths to each possible goal location, as
seen for $C=9$ in Figure~\ref{fig:messages}.
Adding  messages shortens the average path length that the receiver
takes to reach the goal, which  explains why we see such improvements in the
subcapacity regime in Figure~\ref{fig:return}.
In most cases in Figure~\ref{fig:messages} ($C=9$), a single message is used to signal goal
locations lying on the same shortest path, but we also observe cases where
a single message is used to encode a single goal location (e.g, ``7'' and ``3'' in
\emph{Four room}, or ``3'', ``5'' and ``6'' in \emph{Two room}).
In general, as we increase the size of the communication channel
we observe a preference in the sender for assigning a single message to a single location.

\paragraph{Multisender agent setup} 
Message distributions for selected agent setups consisting of two senders, with
two available messages each, are shown in the lower panel in Figure~\ref{fig:messages}.
The panel shows results from five different experiments;  a single experiment
consists of a (Sender \#1, Sender \#2) plot.

We observe a highly coordinated allocation of messages, so that if
the first sender allocates the messages by partitioning the space according to
one axis, the second sender does the same with a different, possibly orthogonal axis. For
example, in the \emph{Pong} case, Sender \#1 partitions the space according to
the \emph{y}-axis, using the message ``0'' to denote ``down'' and ``1'' to
denote ``up''.
Sender \#2 then partitions the space according to the \emph{x}-axis, by assigning message ``1'' to all goal locations to the left of the center,
and ``0'' to all goal locations to the right of the center.
The senders learned to coordinate their
actions even though they are individual agents that only share the information
about the reward $R_t$.

If we consider multiple senders as a single sender emitting several independent messages, we can observe a basic form of compositional structure~\cite{fodor88}.
Thus, an important feature of formal and natural languages~\cite{frege1892} emerges based on minimal assumptions about interactions between individual messages.\footnote{
The expressivity of this protocol is bounded by limiting the number of concatenated messages in a single expression, thus impacting its productivity.}

Lastly, we evaluate the relative impact of individual senders on the task
performance. We investigate whether messages from all senders
contribute equally to the receiver's ability to solve the task.
If all messages are equally important, we expect to
see an equal drop in task performance when we scramble each message
individually.
We test this hypothesis by letting trained agents from the 5S-1R setup perform the task for \numprint{1000}
episodes (with $\epsilon=0$ for both agents), with the following modification: we replace a single message in the
sequence of 5 messages with a random message from that sender's vocabulary. 
This is done iteratively for each message in a single experiment, and we
calculate the average return after each episode.
Thus, we get five returns for one 5S-1R setup that we sort in descending order. 
We use those returns to compute the drop in performance relative to the baseline, which is the performance of trained agents
on \numprint{1000} episodes without any modification to the communication channel.

\begin{figure}[t]
\centering
\includegraphics[width=.4\textwidth]{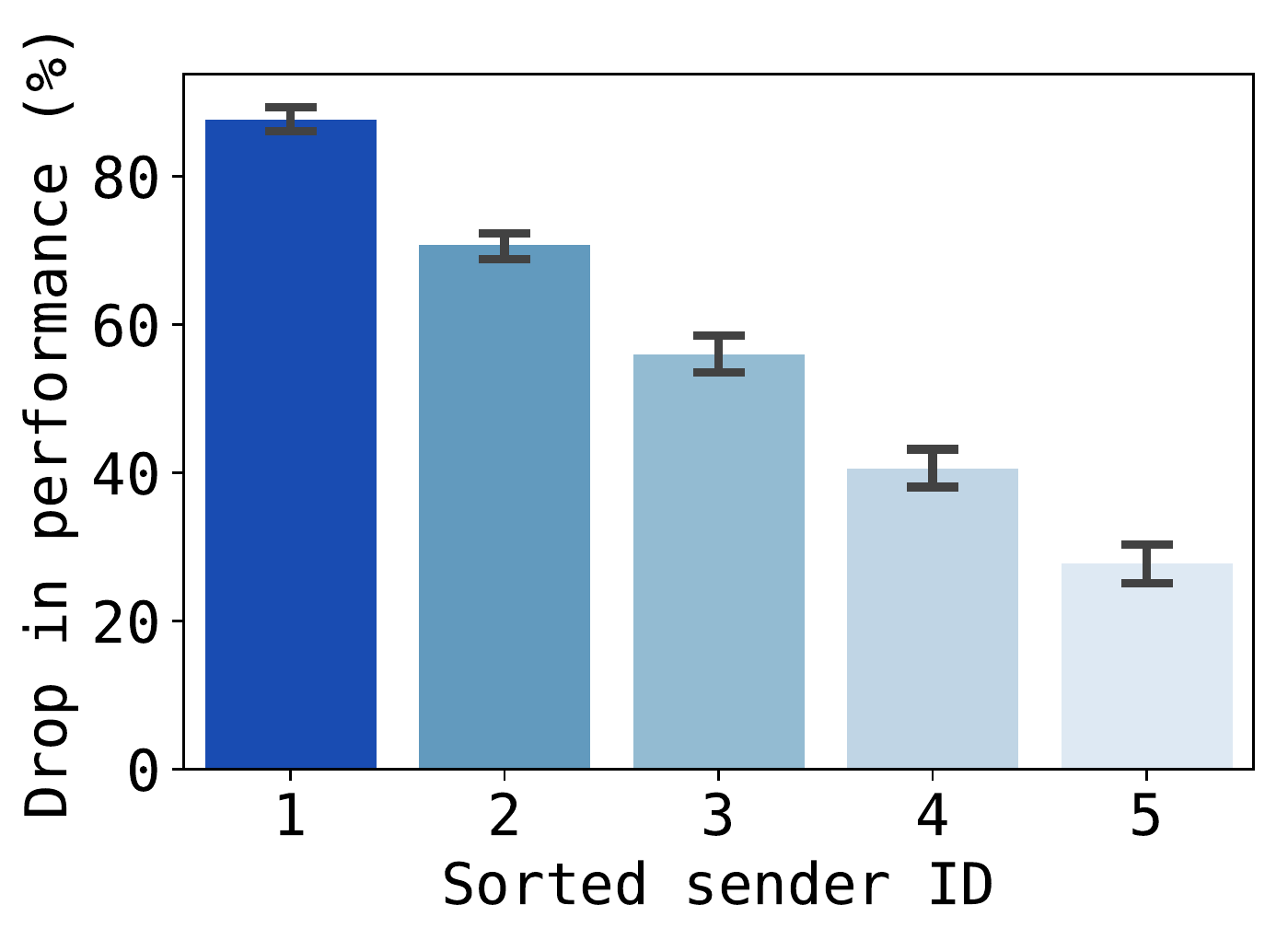}
\caption{Average drop in task performance caused by scrambling each sender's message in
    5S-1R setups. 95\% bootstrapped confidence intervals are shown.}
\label{fig:dominance}
\end{figure}

The results in Figure~\ref{fig:dominance} show the average drop in performance
for all 5S-1R setups.
Since we observe a gradual decrease in performance drop with each subsequent
sender, we can reject the hypothesis that each sender contributes equally.
Scrambling the message from a single sender can cause a performance drop anywhere from 28\% to 88\%.
Thus, while all senders are important for achieving the best performance on
the task, some senders contribute more than the others.
Upon visually inspecting a few sender policies from 5S-1R setups with such dramatic drops, we noticed a pattern in such setups where there is a one ``significant'' sender that allocates one message for the one half of the environment, and another one for the other half.  We speculate this might be an artifact of the training process, though the benefits of it are unclear.

\subsection{The Relationship to Spatial Discourse}
Our results indicate that the characteristics of the communication
protocol adopted by agents depend on the topological layout of the environment,
as well as the communication channel capacity. Despite the contrived nature of
the navigation task used, the emergent protocol exhibits some features that can be meaningfully interpreted
in the context of spatial discourse used in route descriptions. Our results are congruent
with recent research showing that people establish different linguistic strategies when describing locations of goal states in maze environments~\cite{nolle2020}. 
In particular, such strategies are dependent on the environmental structure, landmarks and other features, corroborating the hypothesis that aspects of natural language are shaped by environmental affordances.

Similarly to language used by individuals describing maze environments, emergent signals in our experiments can be interpreted as describing 
lines (e.g. ``Turn left and go down that way'') in stratified environments such
as \emph{Pong}, and paths (``Go 1 step up, 1 right, 1 up'') in more regular
environments such as \emph{Empty room}. However, in our case, these strategies are dependent on the communication channel capacity, as we observe such behaviour only when agents have to compress information about the environment.

\section{Conclusion and Future Work}
Emergent communication in multi-agent reinforcement learning is
a promising avenue for creating AI systems that can learn adaptive communication strategies.
We have shown that situated agents are able to learn
an interpretable, grounded communication protocol that allows them to efficiently,
and in many cases, optimally, solve  navigation tasks in various gridworld
environments. 
By scrutinizing the agents' policies, we have observed that signals such as \emph{left}, \emph{up}, or \emph{upper left room} emerge, in such a way that the state space is clustered spatially with minimal inductive bias.
Using populations of agents to obtain a sequential representation of signals, we have shown that the learned protocol exhibits basic compositional structure as well as signal dominance. Future work will examine those properties and their relationship to natural language in more detail.

\section{Acknowledgments}
The authors would like to thank Angeliki Lazaridou and other colleagues at DeepMind, others at MILA, and Thomas Colin and Terry Stewart for helpful discussions and commentary on this paper. IK's research at University of Waterloo was supported by Ontario Graduate Scholarship.

\bibliographystyle{apacite}

\setlength{\bibleftmargin}{.125in}
\setlength{\bibindent}{-\bibleftmargin}

\bibliography{biblio}

\end{document}